\documentclass[10pt,twocolumn,letterpaper]{article}

\usepackage{iccv}              % To produce the CAMERA-READY version
\definecolor{iccvblue}{rgb}{0.21,0.49,0.74}
\usepackage[pagebackref,breaklinks,colorlinks,allcolors=iccvblue]{hyperref}

\def\paperID{1} % *** Enter the Paper ID here
\def\confName{ICCV HiCV Workshop}
\def\confYear{2025}

\usepackage{graphicx} % Required for inserting images
\usepackage{xcolor}
\usepackage{amsmath}
\usepackage{glossaries}
\usepackage{indentfirst}
\usepackage{booktabs}
\usepackage{multirow}

\newacronym{VLSI}{VLSI}{Very Large Scale Integration}
\newacronym{SHIL}{SHIL}{Subharmonic Injection Locking}
\newacronym{CMOS}{CMOS}{Complementary Metal-Oxide-Semiconductor}
\newacronym{ODE}{ODE}{Ordinary Differential Equation}

\title{OscNet v1.5: Hopfield Network Designed for Image Classification by Energy-Efficient Oscillator Fabrics}

\author{Wenxiao Cai$^{1}$\thanks{Wenxiao Cai, Zongru Li and Iris Wang contributed equally to this paper. This work was done when Iris Wang was visiting Stanford University. Correspond to: Thomas H. Lee: tomlee@ee.stanford.edu and Wenxiao Cai, wxcai@stanford.edu.} \hspace{1em}Zongru Li$^{1}$\footnotemark[1]\hspace{1em} Iris Wang$^{2}$\footnotemark[1] \hspace{1em} Yu-Neng Wang$^{1}$\hspace{1em}Thomas H. Lee$^{1}$\ \\ \\
$^1$ Stanford University\hspace{1em}
$^2$ Carnegie Mellon University
}

\begin{document}

\maketitle

\begin{abstract}
Machine learning has achieved remarkable advancements but at the cost of significant computational resources. 
This has created an urgent need for a novel and energy-efficient computational fabric and corresponding algorithms.
CMOS Oscillator Networks (OscNet) is a brain inspired and specially designed hardware for low energy consumption.
In this paper, we propose a Hopfield Network based machine learning algorithm that can be implemented on OscNet.
The network is trained using forward propagation alone to learn sparsely connected weights, yet achieves an $8\%$ improvement in accuracy compared to conventional deep learning models on MNIST dataset.  
OscNet v1.5 achieves competitive accuracy on MNIST and is well-suited for implementation using CMOS-compatible ring oscillator arrays with \gls{SHIL}.
In oscillator-based inference, we utilize only $24\%$ of the connections used in a fully connected Hopfield network, with merely a $0.1\%$ drop in accuracy.
OscNet v1.5 relies solely on forward propagation and employs sparse connections, making it an energy-efficient machine learning pipeline designed for oscillator computing fabric. 
The repository for OscNet family is: \url{https://github.com/RussRobin/OscNet}. 
\end{abstract}

\section{Introduction}
\begin{figure}[h]
	\begin{center}
    \includegraphics[width=1\linewidth]{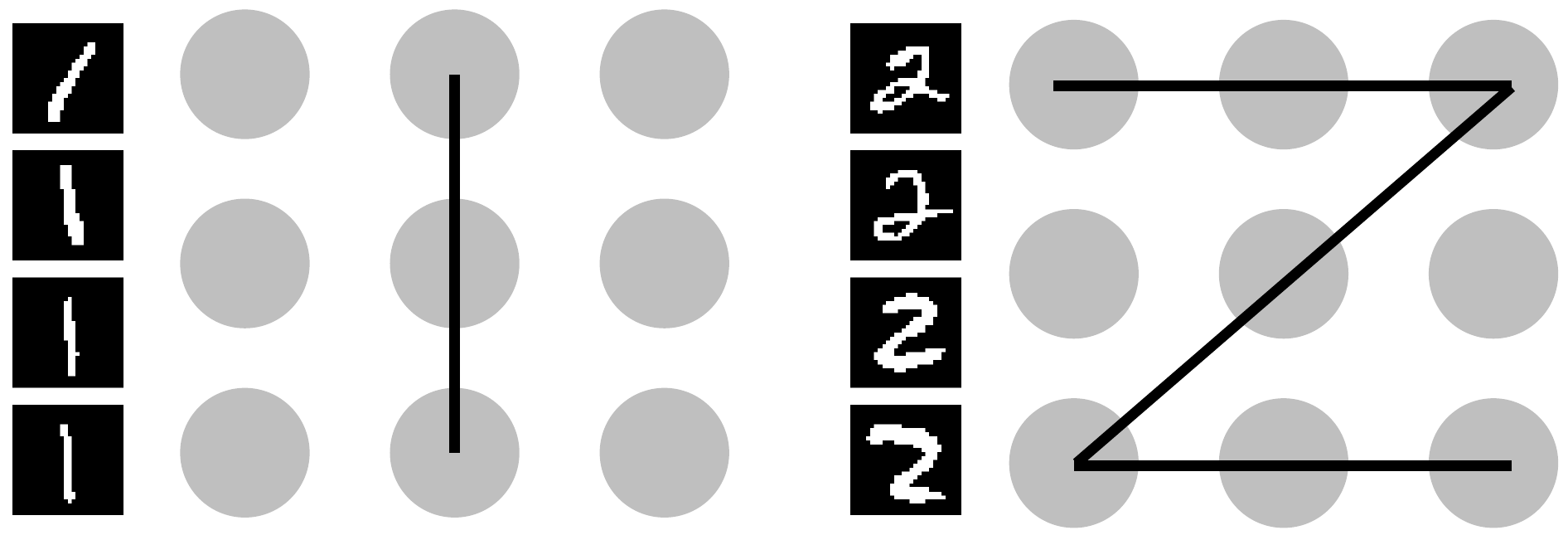}
	\end{center}
\caption{A simplified illustration of sparsely connected Hopfield Network for image classification. Circle: node in Hopfield Network. Line: connections between nodes. Smaller figures: training images.}
\label{fig:teaser}
\end{figure}
Gradient descent based machine learning has achieved significant success in various fields~\cite{spatialbot,ml_finance_1,ml_health_1}. 
However, the energy consumption of current von Neumann architecture based CPU and GPU computing is becoming increasingly high and unsustainable~\cite{train_cost_1,train_cost_2,train_cost_3,train_cost_4}.
In contrast, the human brain can perform complex tasks under extremely low power, even though the efficiency of neural signal transmission is far lower than that of electric currents in CMOS circuits. 
Inspired by this, the Complementary Metal-Oxide-Semiconductor (CMOS) Oscillator Network (OscNet) was developed, where information is directly stored in the phase of each oscillator, and inter-oscillator currents model communication between neurons.
This architecture is especially efficient on forward propagation in machine learning.
% Prior work has designed Hebbian learning schemes~\cite{oscnet} and specific algorithms for solving nondeterministic polynomial-time (NP) problems~\cite{graph_coloring} on OscNet.
Prior work~\cite{graph_coloring} developed a mathematical model to optimize graph coloring using oscillator circuits, and~\cite{oscnet} designed Hebbian learning schemes on OscNet.
Building on this foundation, we propose OscNet v1.5, a brain-inspired and energy-efficient Hopfield Network–style algorithm to support learning-based generalization from original Hopfield Network to image classification tasks~\cite{anitha2021hyperbolic, abernot2023training}, designed for the oscillator computing fabric.

A simplified illustration of the proposed algorithm is shown in Fig.~\ref{fig:teaser}.
We develop a biologically plausible pipeline for Hopfield-based learning and prediction that relies solely on local information and online learning, consistent with the constraints observed in real neural systems.
Training of OscNet v1.5 is based on forward propagation only~\cite{hebb1,hebb2,oscnet}, so it is more energy efficient then current deep learning networks which rely heavily on forward and backward propagation.

Traditional Hopfield Networks are fully connected, which does not reflect the connectivity structure observed in the human brain. In this paper, we construct sparsely connected Hopfield Networks. 
Our sparsely connected architecture can outperform fully connected ones in classification tasks—in terms of both accuracy and computational efficiency—while also being more biologically plausible. 
We propose a Hebbian style learning rule that determines not only the connection weights but also whether a connection should exist between nodes. 
In oscillator based inference, we use only $24\%$ of connections while the performance only drops by $0.1\%$ compared to a fully connected Hopfield Network.
Our algorithm is energy efficient because it learns from forward propagation only, without the need to do back propagation or gradient descent. Additionally, it is a sparsely connected network, which further conserves energy.
The main contributions of this paper are:
\begin{itemize}
    \item We proposed OscNet v1.5, a machine learning pipeline designed for oscillator computing fabric. OscNet v1.5 is energy efficient as it relies solely on forward propagation and employs sparse connections.
    \item OscNet v1.5 reaches $8\%$ improvement in CPU-based classification performance compared to forward- and backward-propagation-based deep learning models on MNIST~\cite{mnist} dataset.
    \item In oscillator-based inference, we utilize only $24\%$ of the connections used in a fully connected Hopfield network, with merely a $0.1\%$ drop in accuracy.
\end{itemize}

\section{Background}

\subsection{Traditional Hopfield Networks}
The Hopfield Network~\cite{hopfield, cheng1996application} is widely regarded as a potential computational model for understanding biological memory mechanisms. Based on energy minimization, Hopfield Networks can store specific patterns as attractor states. Even when presented with noisy or partial inputs, the network can reliably converge to the closest stored pattern with Hebbian learning, mimicking aspects of human associative recall.
However, biological memory systems go far beyond this static recall ability. In real organisms, memory is not limited to recalling fixed patterns. It also involves learning to generalize from experience. 
For example, when we learn to recognize the digits 0 through 9, we encounter a wide variety of image inputs. Simply memorizing a few training examples, as in a standard Hopfield Network, does not capture the flexibility and robustness of human perception and memory.
A Hopfield Network consists of \( N \) neurons, each of which is fully connected to every other neuron in the network but not to itself. 
The connections are symmetric, meaning the weight matrix \( \mathbf{W} \) satisfies:
\begin{equation}
    w_{ij} = w_{ji}, \quad w_{ii} = 0
\end{equation}
The state of each neuron \( i \) is represented by a binary value:
\begin{equation}
s_i \in \{-1, 1\}
\end{equation}

The dynamics of a Hopfield Network involve asynchronous updates of neurons to restore its saved pattern. The update rule for neuron \( i \) is:
\begin{equation}
s_i^{\text{new}} = \text{sgn}\left(\sum_{j} w_{ij} s_j - \theta_i\right)
\end{equation}
where \( \theta_i \) is the threshold for neuron \( i \).
The network evolves to minimize an energy function defined as:
\begin{equation}
E(\mathbf{s}) = -\frac{1}{2} \sum_{i,j} w_{ij} s_i s_j + \sum_i \theta_i s_i
\end{equation}
Hopfield Networks can store patterns by adjusting the weights using Hebbian learning:
\begin{equation}
w_{ij} = \frac{1}{N} \sum_{\mu=1}^P \xi_i^\mu \xi_j^\mu
\end{equation}
where \( P \) is the number of patterns \( \boldsymbol{\xi}^\mu \) to be stored.

\subsection{CMOS Oscillator Networks}
In this section, we review the mathematical theory established in~\cite{graph_coloring}, upon which OscNet~\cite{oscnet} builds. CMOS Oscillator Network (OscNet)~\cite{oscnet} operates with high-order injection locking, with local connections to neighboring oscillators. 
Each oscillator is injected with a master pump signal at $N·f_0$, where $f_0$ is oscillators' natural oscillator frequency, and $N$ is a chosen integral.
The phases of oscillators can be $n \frac{2\pi}{N}$, where $n=1,2...N-1$.
This network mimics the activity of polychronous spiking neurons in the brain~\cite{oscillator_spiking1}, where each oscillator represents a single brain cell. 
In OscNet, all oscillators oscillate at the same frequency, and the information is encoded in the phases of the individual oscillators.  The voltage across the oscillator $i$ at time $t$ is:
\begin{equation}
    V_i(t)=A_i cos(\omega_0t+\theta_i),
\end{equation}
where $\omega_0$ is the free-running frequency of the oscillator, $A_i$ is the amplitude, and $\theta_i$ is the phase.
% The system can be viewed as oscillators interacting with their neighbors and the main pump, with current serving as the medium of interaction. 
Each oscillator outputs a current based on its own state, while changes in its phase are determined by the influence of neighboring oscillators and the main pump, which can be explained by Impulse Sensitivity Function (ISF) theory~\cite{isf_1}.
Kuramoto oscillators are a class of coupled phase oscillators that model synchronization phenomena in biological and physical systems. Unlike the binary states in Hopfield networks, Kuramoto models operate in continuous phase space, making them well-suited for analog implementations.
The phase \(\theta_i\) of each oscillator evolves according to:~\cite{kuramoto_eq,graph_coloring}:
\begin{equation}
    \frac{d\theta_i}{dt} = \sum_{(i,j) \in \mathcal{N}} K_{ij} sin(\theta_i - \theta_j) + K_p sin(N\theta_i),
\end{equation}
where $K_{i,j}$ represents the coupling strength between oscillator $i$ and $j$,  $K_p$ models the main pump current and ISF, and $\mathcal{N}$ is the set of neighboring oscillators of $i$.
In OscNet v1.5, we modify this equation to incorporate Hebbian-trained weights and \gls{SHIL}, enabling phase-based computation for both memory retrieval and image classification. This oscillator-based model unifies energy minimization, biological plausibility, and hardware feasibility within a single computational framework.
The global Lyapunov function exists and will be minimized over time by the oscillator network~\cite{graph_coloring,oscnet}: 
\begin{equation}
    E(t) = \frac{N}{2} \sum_{(i,j) \in  \mathcal{N}} K_{ij} cos(\theta_i - \theta_j) + \sum_{i} K_p cos(N\theta_i).
    \label{eq:Lyapunov}
\end{equation}

Eq.~\ref{eq:Lyapunov} takes a similar form of Potts Hamiltonian~\cite{potts_1}:
\begin{equation}
    H = - \sum_{(i,j) \in \mathcal{N}} J_{ij} cos(\theta_i - \theta_j) - \sum_{i} h_i.
    \label{eq:potts}
\end{equation}

Potts Hamiltonian was initially developed to model overall energy and phase transitions in ferromagnetic materials.
The CMOS oscillator network can find the minimal value of Potts Hamiltonian and thus minimize energy of the system. In this paper, we model problems with Potts Hamiltonian, design structures of OscNet and it finds optimal phases of the system.

\subsection{Problem Definition}
We follow the concept of traditional Hopfield Networks, assuming that memories are stored in the connection weights of a Hopfield-like network. 
We further hypothesize that certain regions of the human brain operate as Hopfield Networks. 
For simplicity, we assume that each Hopfield Network stores only one pattern, and the brain employs multiple Hopfield Networks to store different patterns.
When we perform visual recognition—for example, distinguishing the digit “1” from “2”—the brain compares the perceived pattern to the patterns stored in these networks. 
Specifically, it loads the observed pattern into each Hopfield Network and identifies which network minimizes its energy function, so as to determine which stored pattern is closest to the current observation. This serves as a mechanism for recognition and classification.
The problem we are interested is: how does the brain construct the connection weights of these Hopfield Networks?

\section{Proposed Method}
\subsection{Pipeline}
In OscNet v1.5, each node in Hopfield Network corresponds to one pixel of an image. 
We assume all input images have the same resolution and that each pixel is binarized. Let $p_i \in \{0,1\}$ denote the $i$th pixel of an image in the training set $D$, which contains images from classes $C_{1},\dots,C_{n}$. 
To match Hopfield conventions, we map $0 \mapsto -1,\quad 1 \mapsto +1$ in accordance with Hopfield annotations.
We allocate one Hopfield network per class.
For each training image in class $C_k$, we update the symmetric weight matrix $W^{(k)} = [w_{ij}^{(k)}]$ by Hebbian learning. Starting from $W^{(k)} = 0$, each pixel pair $\{p_i,p_j\}$ contributes:
\begin{equation}
  w_{ij}^{(k)} \;=\; w_{ij}^{(k)} + p_ip_j.
\end{equation}
Thus, if $p_i$ is always equal to  $p_j$ across many images of in a class $C_k$, then \(w_{ij}^{(k)}\) becomes large and positive.
If $p_i=-p_j$ is true across many images of class $C_k$, then $w_{ij}^{(k)}$ becomes large in absolute value and negative.
Large $\lvert w_{jk}^{(i)}\rvert$ indicates a strong correlation or anti-correlation between pixels \(j\) and \(k\), which helps discriminate class \(C_i\).
In the test time, given a test image with pixels \(p'_j\in\{-1,+1\}\), we compute its energy under each class:
\begin{equation}
  E_k = \sum_{ij} w_{ij}^{(k)}\,p'_i\,p'_j.
\end{equation}
If the sign of \(p'_j p'_k\) agrees with that of \(w_{jk}^{(i)}\), it lowers the energy, indicating similarity to class $C_k$. The test image is loaded to each Hopfield Network, and assigned to the class whose Hopfield Network yields the lowest energy.

\subsection{Sparse Connection}
\begin{figure}[h!]
	\begin{center}
    \includegraphics[width=1\linewidth]{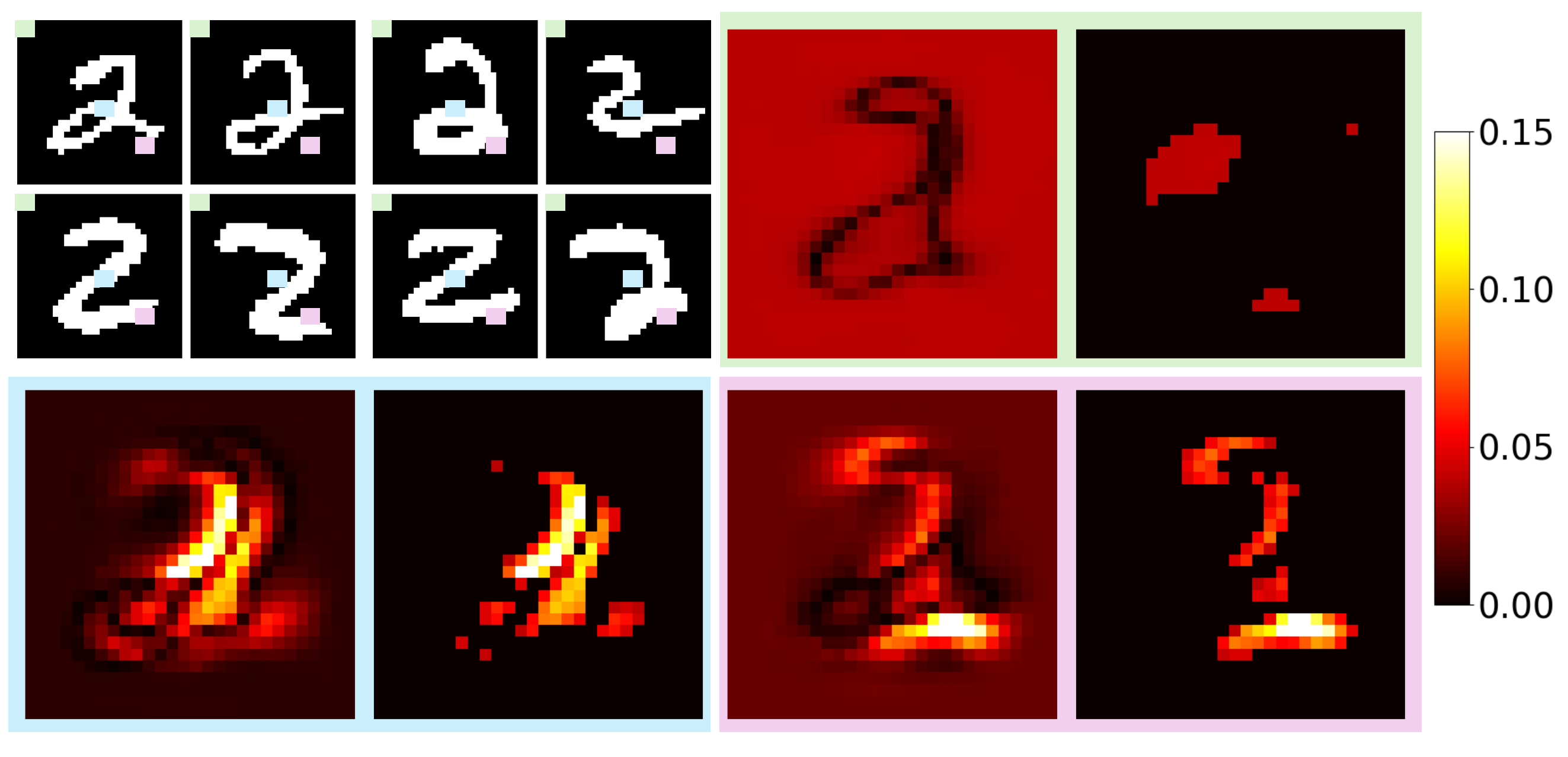}
	\end{center}
\caption{Features learned on three points of the dataset, as shown with green, blue and purple points. Top left: 8 training images. Remaining: full connection (left) and selected sparse connections of the three points. The weights shown have been normalized across class and pixel channel.}
\label{fig:sparse_2}
\end{figure}

The traditional Hopfield network is fully connected. However, in OscNet v1.5, connections with small 
$w_{ij}^{(k)}$ convey very little information. Therefore, we prune these connections and retains only the $N_{1}$ largest weights. To ensure spatial coverage, we also require that each pixel maintain at least $N_{2}$ connections. If each image has $P$ pixels and we set $N_{1} = P\,N_{2}$, then for each pixel $i$ we keep its top $N_{2}$ most informative connections rather than selecting the global top $N_{1}$. Note that although the underlying weight matrix is symmetric, $w_{ij} = w_{ji}$, the resulting sparse connectivity pattern need not be symmetric.
For a given test sample, the pixel values of two points may not always be the same due to random noise. The sparse connection filtering removes the connections with low information content.
Selected sparse connections are shown in  Fig.~\ref{fig:sparse_2}. Sparse connections in OscNet v1.5 can be viewed as a noise filter: we retain only the pixel pairs that provide substantial information relevant to the image classification.

\subsection{Normalization}
\begin{figure}[h!]
	\begin{center}
    \includegraphics[width=1\linewidth]{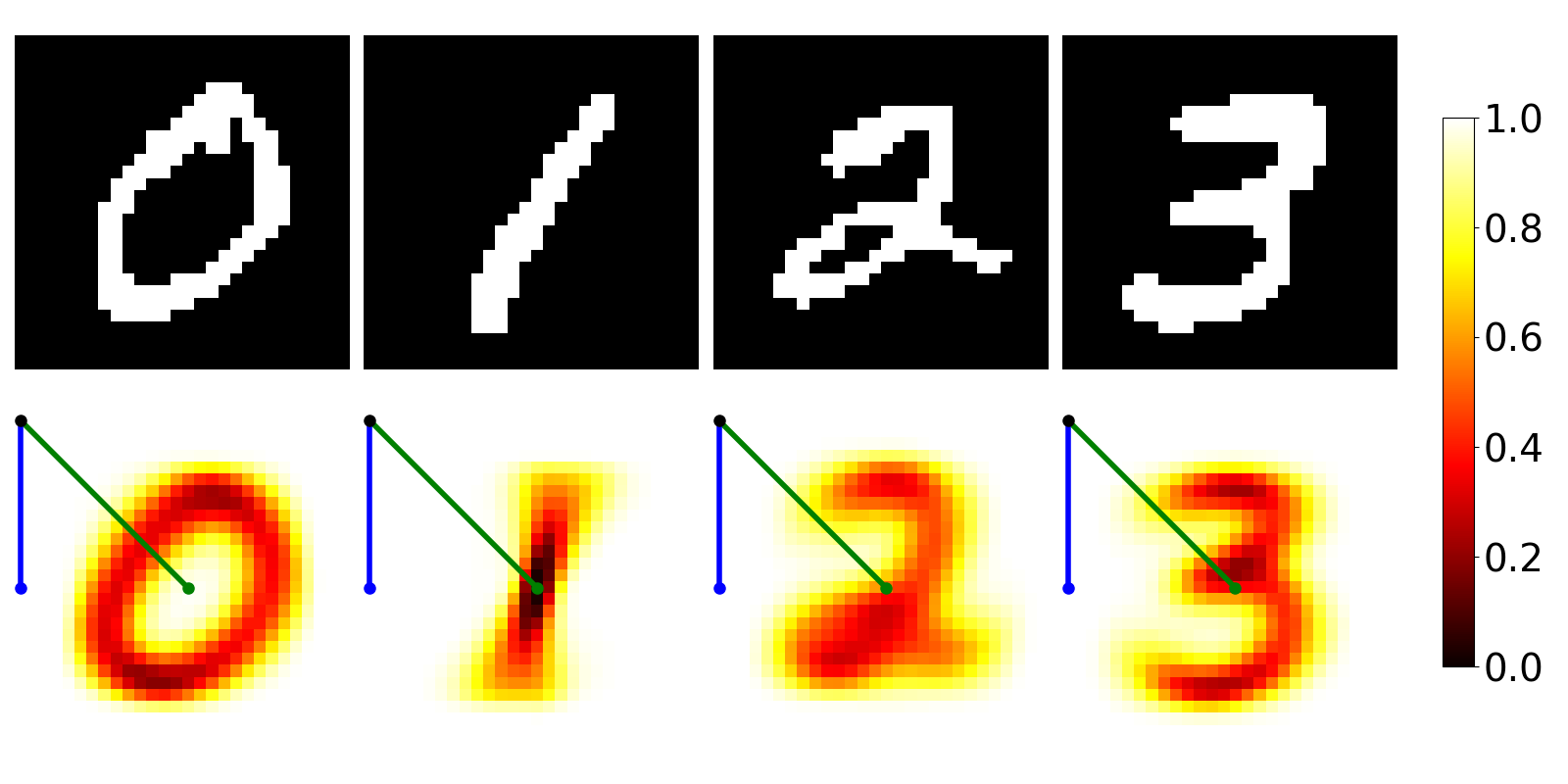}
	\end{center}
\caption{Up: examples of training images of class 0,1,2,3 in MNIST~\cite{mnist}. Down: learned features of pixel (0,0) in Hopfield Networks for class 0,1,2,3. The colormap shows absolute value of similarity matrix. Black: pixel (0,0), green: pixel (14,14), and blue: pixel (0,14). The similarity of (0,0)-(0,14), as shown with blue line, is always large across labels, which brings little information in image classification. (0,0)-(14,14) connection in green tells between class 0 and 1. The weights shown are not normalized.}
\label{fig:dim0_norm}
\end{figure}

After calculating the entropy of each connection, we obtain a similarity matrix of size $n \times P \times P$, where $n$ is the number of classes and $P$ is the number of pixels. 
This matrix is then used to compute the energy of test images in the $n$ Hopfield Networks. However, if in each class of images the values of $|w_{ij}|$ are all large, then these connections do not specifically indicate any particular class of images. 
As shown in Fig.~\ref{fig:dim0_norm}, for example, the pixel pair $(0,0)$ and $(14,14)$ can provide a significant amount of information in class 0: if these two pixels in the test image have the same value, it is very likely that the image corresponds to class 0. However, the same pixel pair does not provide much useful information in class 1.
If a pixel $i$ has a large similarity only with a few other pixels and low similarity with the rest, the connections with those few pixels carry more informative content. On the contrary, if a pixel has a large similarity with many other pixels, the information it carries is relatively low.
Therefore, we normalize the similarity values so that the magnitude of similarity reflects the amount of information a particular connection provides for distinguishing between different classes.

\subsection{Kuramoto-Based Hopfield Inference}

OscNet v1.5 replaces the traditional discrete-state dynamics with continuous-time oscillator evolution governed by a modified Kuramoto model. Each image is encoded into a phase vector \( \theta_i(t) \), where binarized pixel values map as \( 0 \rightarrow \pi \) and \( 1 \rightarrow 0 \). The system evolves under~\cite{kuramoto_eq,wang2024shem}:
\begin{equation}
\frac{d\theta_i}{dt} = -\epsilon_i \sin(2\theta_i) - \sum_j w_{ij} \sin(\theta_i - \theta_j),
\end{equation}
where \( \epsilon_i \) represents the strength of \gls{SHIL}, which stabilizes oscillator phases to binary values (0 or \( \pi \)), and \( w_{ij} \) is the learned coupling strength.

The final spin state is decoded via:
\begin{equation}
s_i = \text{sign}[\cos(\theta_i)] \in \{-1, +1\}.
\end{equation}
Energy is then computed in the Ising form~\cite{mohseni2022ising}:
\begin{equation}
E = -\sum_{i,j} w_{ij} s_i s_j.
\end{equation}

Classification proceeds by simulating the oscillator network using numerical \gls{ODE} solvers (e.g., Runge-Kutta), evaluating the energy for each class, and selecting the one with minimum energy.
This phase-based mechanism enables a smooth and differentiable dynamics that converges naturally to attractor states. When initialized with noisy images, the network simultaneously denoises and classifies the input. Sparse connectivity ensures energy efficiency and biological plausibility.

\begin{figure}[h!]
	\begin{center}
    \includegraphics[width=1\linewidth]{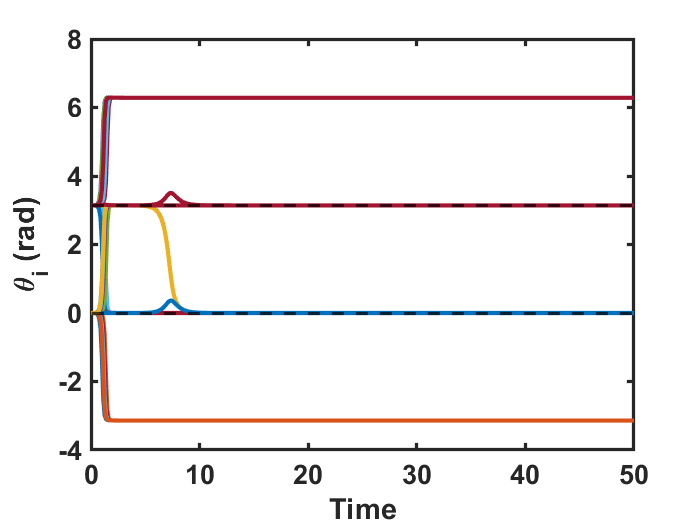}
	\end{center}
\caption{Phase trajectories of selected oscillators. The input is a noisy digit 0; weights are trained on digits 0 and 1.}
\label{fig:phase}
\end{figure}

\begin{figure}[h!]
	\begin{center}
    \includegraphics[width=1\linewidth]{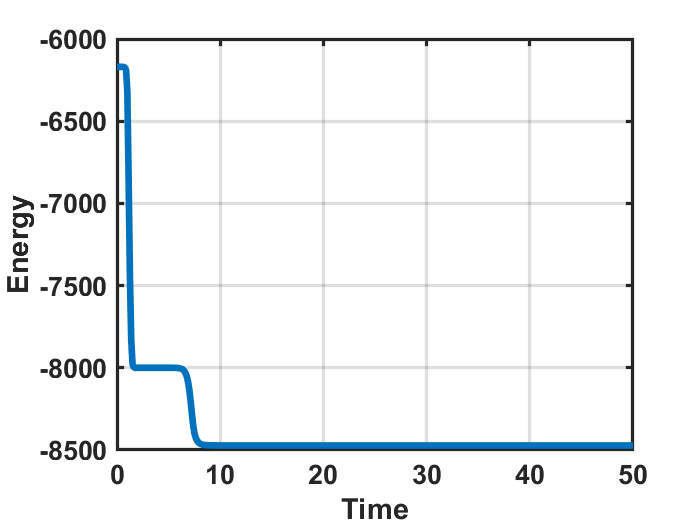}
	\end{center}
\caption{Energy evolution during convergence. The system rapidly minimizes energy and stabilizes to a memorized state.}
\label{fig:energy}
\end{figure}

\begin{figure}[h!]
	\begin{center}
    \includegraphics[width=1\linewidth]{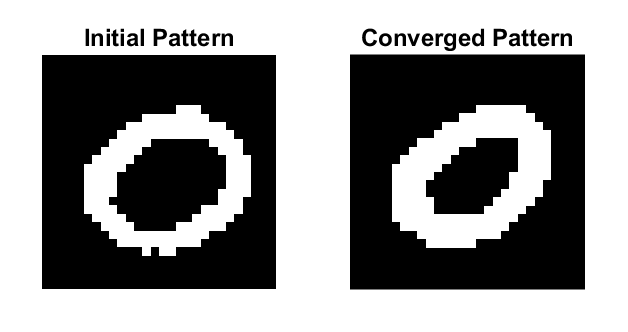}
	\end{center}
\caption{Left: Input image (noisy digit 0). Right: Reconstructed image after convergence. The network restores the correct pattern.}
\label{fig:image_recover}
\end{figure}

Figure~\ref{fig:phase} shows a representative phase trajectory when the network is initialized with a noisy digit 0 image, using weights trained on digits 0 and 1. The oscillators converge to a clean phase-locked pattern corresponding to one of the stored digits. Figure~\ref{fig:energy} displays the corresponding energy evolution, illustrating rapid convergence to a low-energy state. Figure~\ref{fig:image_recover} shows the input image and the reconstructed output after convergence. It is also important to note that classical Hopfield networks have limited storage capacity and typically cannot stably store more than a few uncorrelated patterns~\cite{belyaev2020classification}. In this example, the weight matrix is trained on just two patterns, digits 0 and 1, so the network can accurately recover the correct pattern when the input is a noisy version of either digit. However, if the input does not resemble either stored pattern (e.g., a different digit), the network may still converge to one of the two attractor states arbitrarily.
We use the learned weights in OscNet v1.5 to select connections in oscillator Hopfield Network. We set connections to $0$ if the weights are close or equal to $0$.

\section{Experiments}

\begin{table}[h]
\caption{Binary MNIST~\cite{mnist} results of two layer \textbf{nonlinear} classifier (FC-ReLu-FC, FC: fully connected layer) with image input and OscNet v1.5 representation of images as inputs. OscNet v1.5 represents images as similarity matrices. $\#$ Hidden is the number of hidden layers between two FC layers.
}
\centering

\begin{tabular}{lcc}

\toprule
Input Feature &  $\#$ Hidden & Accuracy ($\%$)\\
\hline
Image & \multirow{2}{*}{10} & 89.30\\
\textbf{Ours} &  & \textbf{96.23}\\
\hline
Image & \multirow{2}{*}{16} & 92.45\\
\textbf{Ours} &  & \textbf{96.35} \\
\hline
Image & \multirow{2}{*}{64} & 95.38\\
\textbf{Ours} &  & \textbf{97.35} \\
\hline
Image & \multirow{2}{*}{128} & 96.83\\
\textbf{Ours} &  & \textbf{97.40} \\

\bottomrule
\end{tabular}
% \vspace{-1em}
\label{tab_nonlinear_exp}
\end{table}

\begin{table}[h]
\caption{OscNet v1.5 performance on binary MNIST~\cite{mnist} image classification dataset with different number of connections. With $n\times784$ connections, Acc-Top uses top $\frac{n}{2}$ similarities closest to 1 and $\frac{n}{2}$ similarities closest to -1 for each pixel. Acc-TopABS chooses $n$ largest absolute values for sparse connection. The best performance is marked with \textbf{bold} text and second best is \underline{underlined}.}
\centering

\begin{tabular}{lcc}

\toprule
$\#$ Connections & Acc-Top ($\%$) & Acc-TopABS ($\%$) \\
\hline
fully connected & \multicolumn{2}{c}{78.65} \\  
\hline
10 & 78.27 & 82.46 \\
16 & 81.13 & 84.30\\
28 & 83.73 & 85.86\\
56 & 85.86 & \textbf{86.55} \\
64 & 85.97 & 86.15 \\
84 & \underline{86.12} & \underline{86.32} \\
128 & \textbf{86.15} & 83.90 \\
256 & 83.89 & 79.61 \\

\bottomrule
\end{tabular}
% \vspace{-1em}
\label{tab_sparse_exp}
\end{table}

\begin{table}[h]
\caption{Experimental results of Hebbian learning~\cite{oscnet}, Autoencoder based Backprop and OscNet v1.5, finetuned by fully connected \textbf{linear} classifiers, on binary MNIST~\cite{mnist}. $\#$ Hidden for OscNet v1.5 stands for number of non-zero connections.}
\centering

\begin{tabular}{lcc}

\toprule
Model &  $\#$ Hidden & Accuracy \\
\hline
Hebbian & \multirow{3}{*}{10} &  77.47 \\
BackProp &  & 73.84 \\
\textbf{Ours} &  & \textbf{95.44}\\
\hline
Hebbian & \multirow{3}{*}{16} & 85.61  \\
BackProp &  &  81.22 \\
\textbf{Ours} &  & \textbf{96.03} \\
\hline
Hebbian & \multirow{3}{*}{64} & 90.58 \\
BackProp &  & 85.61 \\
\textbf{Ours} &  & \textbf{96.21} \\
\hline
Hebbian & \multirow{3}{*}{128} & 91.33 \\
BackProp &  & 91.35 \\
\textbf{Ours} &  & \textbf{96.08} \\

\bottomrule
\end{tabular}
% \vspace{-1em}
\label{tab_linear_exp}
\end{table}

\begin{table}[h]
\caption{Ablation of normalization over class dimension (Dim 1) and pixel dimension (Dim 2) on binary MNIST~\cite{mnist}. We directly use OscNet v1.5 without finetuing with linear classifier. Our model has $84\times784$ non-zero sparse connections.}
\centering

\begin{tabular}{lccccc}

\toprule
None & Dim 1 & Dim 2 & Dim 1,2   \\
\hline

15.49 & 78.40 & 33.32 & \textbf{86.32}\\

\bottomrule
\end{tabular}
% \vspace{-1em}
\label{tab_norm}
\end{table}

First, we show that Hopfield Network representation of images in OscNet v1.5 is better than original results. 
We use the similarity learned by our model and feed it into 2 layers of fully connected linear classifier. 
The nonlieanr classifier performs better on OscNet v1.5 features than on original images, as shown in Table.~\ref{tab_nonlinear_exp}.
We compare performance of our model with sparse and full connection in Table.~\ref{tab_sparse_exp}.
OscNet v1.5 performs better on sparse connection. 
We suppose that retaining only the larger similarities is equivalent to filtering out the noise caused by similarity connections with less information.
We deploy two models as baselines on MNIST~\cite{mnist}. Hebbian learning~\cite{oscnet} is a biologically inspired training strategy based on forward propagation only. BackProp stands for Autoencoder trained with forward and backward propagation. Features learned by Hebbian learning and BackProp are finetuned with a fully connected linear classifier. Our model does not rely on backward propagation to update weights. $\#$ Hidden stands for number of hidden layers. OscNet v1.5 is sparsely connected and the total number of connections is $784\times\#$ Hidden, the same as Hebbian and BackProp feature extraction. The linear classification results are shown in Table.~\ref{tab_linear_exp}.
OscNet v1.5 performs better than Hebbian learning and forward-backward propagation based Autoencoder.
We do ablation studies on the choice of large similarity connections: choose $n$ connections with largest absolute values, or $\frac{n}{2}$ similarities closest to 1 and $\frac{n}{2}$ similarities closest to -1 for each pixel. Results in Table.~\ref{tab_sparse_exp} show that there is not much difference between these two choices.
We also experiment with the normalization strategies in OscNet v1.5, and the results are shown in Table.~\ref{tab_norm}. Normalizing over both class and pixel dimension leads to highest performance.
The OscNet v1.5 pipeline, along with CMOS oscillator computing, are a new and promising computational fabric. 

\subsection{Pairwise Elimination with Kuramoto Dynamics}

To extend classification from two digits to all 10 MNIST~\cite{mnist} digits, we implement a tournament-style pairwise elimination strategy using oscillator dynamics. For each digit pair \( (i, j) \), we construct a class-specific weight matrix:
\begin{equation}
W^{(i,j)} = \xi^{(i)} (\xi^{(i)})^T + \xi^{(j)} (\xi^{(j)})^T,
\end{equation}
where \( \xi^{(i)} \) is the prototype pattern for digit \( i \). During inference, the input image is mapped to a phase vector \( \theta(t) \) and evolved under the Kuramoto oscillator dynamics parameterized by \( W^{(i,j)} \).

Each pairwise match selects the class whose prototype has the higher cosine similarity to the final oscillator phases. This cascaded decision process decomposes the 10-class classification into a sequence of binary comparisons, enabling scalability with minimal hardware logic and high parallelizability. Although this method is suboptimal compared to global multi-class optimization, it is biologically inspired and highly hardware-efficient.

\begin{figure*}[h!]
\centering
\includegraphics[width=0.7\linewidth]{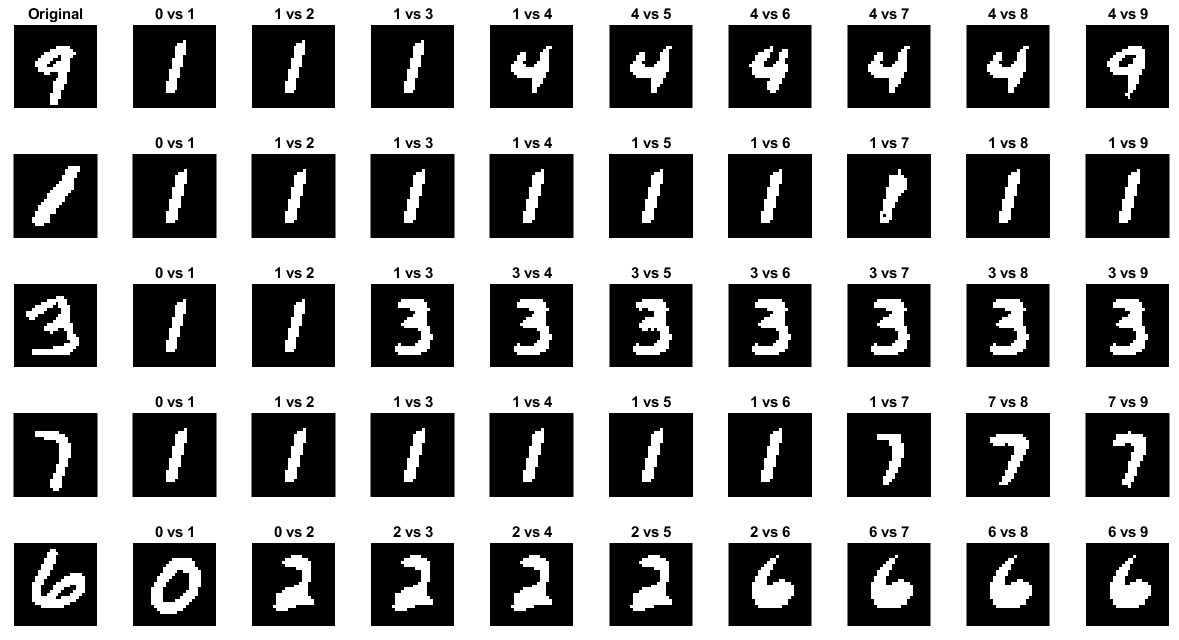}
\caption{Pairwise classification results for five MNIST test digits using Kuramoto dynamics. The first column shows the original image. Each subsequent column shows the output of a binary comparison between the test digit and a different class prototype.}
\label{fig:pairwise_grid}
\end{figure*}

\begin{figure}[h!]
\centering
\includegraphics[width=1\linewidth]{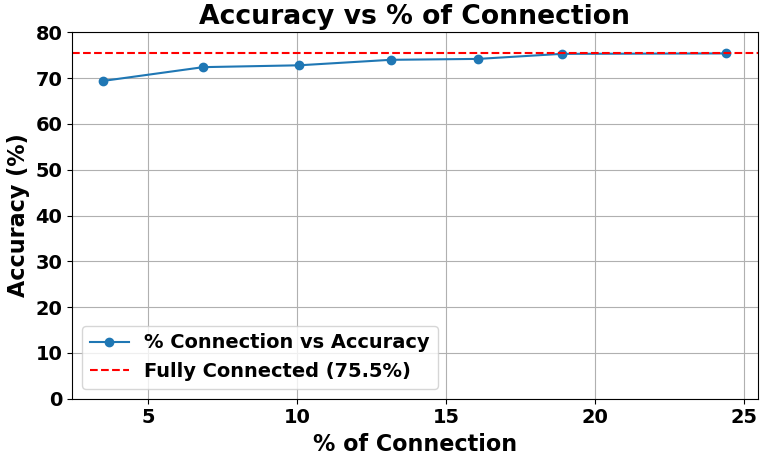}
\caption{Binary MNIST image classification accuracy of oscillator Hopfield Networks with different percentage of sparse connections.}
\label{fig:hopfield_sparse_acc}
\end{figure}

Figure~\ref{fig:pairwise_grid} visualizes the pairwise decision process for five sample MNIST~\cite{mnist} test digits. Each row corresponds to a test digit, and each column shows the binary classification outcome against a different digit prototype. The results show that the input digits are correctly classified in most cases, demonstrating the feasibility of oscillator-based multi-class classification.
Fig.~\ref{fig:hopfield_sparse_acc} presents the image classification accuracy of Hopfield Networks under different levels of sparse connectivity. 
The sparsity patterns are guided by the learned weights from OscNet, which promotes energy-efficient design in oscillator-based Hopfield Networks. 
Using only $24\%$ of the connections, we achieve an accuracy of $75.3\%$, merely $0.1\%$ lower than the fully connected version, while significantly reducing energy consumption. 
Even with just $13\%$ of the connections, OscNet v1.5 incurs only a $1.5\%$ drop in accuracy, offering a favorable trade-off between energy efficiency and classification performance.

\section{Hardware Benefits}

The proposed Kuramoto-based OscNet v1.5 is designed with hardware feasibility in mind. Its continuous-time dynamics and sparse connectivity make it particularly well-suited for analog \gls{CMOS} implementations.

A promising candidate for physical realization is a network of compact ring oscillators, where each oscillator's phase encodes the state of a neuron. Binary quantization of these phases can be enforced using \gls{SHIL}, enabling convergence to discrete states (e.g., \( 0 \) or \( \pi \)) corresponding to Ising spins. The inherent phase evolution in these oscillators closely approximates Kuramoto dynamics, while the learned coupling weights \( w_{ij} \) can be implemented using programmable resistive elements or digitally controlled transmission gates. Oscillator phases can be sampled via pseudo-analog phase detectors and then binarized for classification.

\begin{figure}[h!]
    \centering
    \includegraphics[width=1\linewidth]{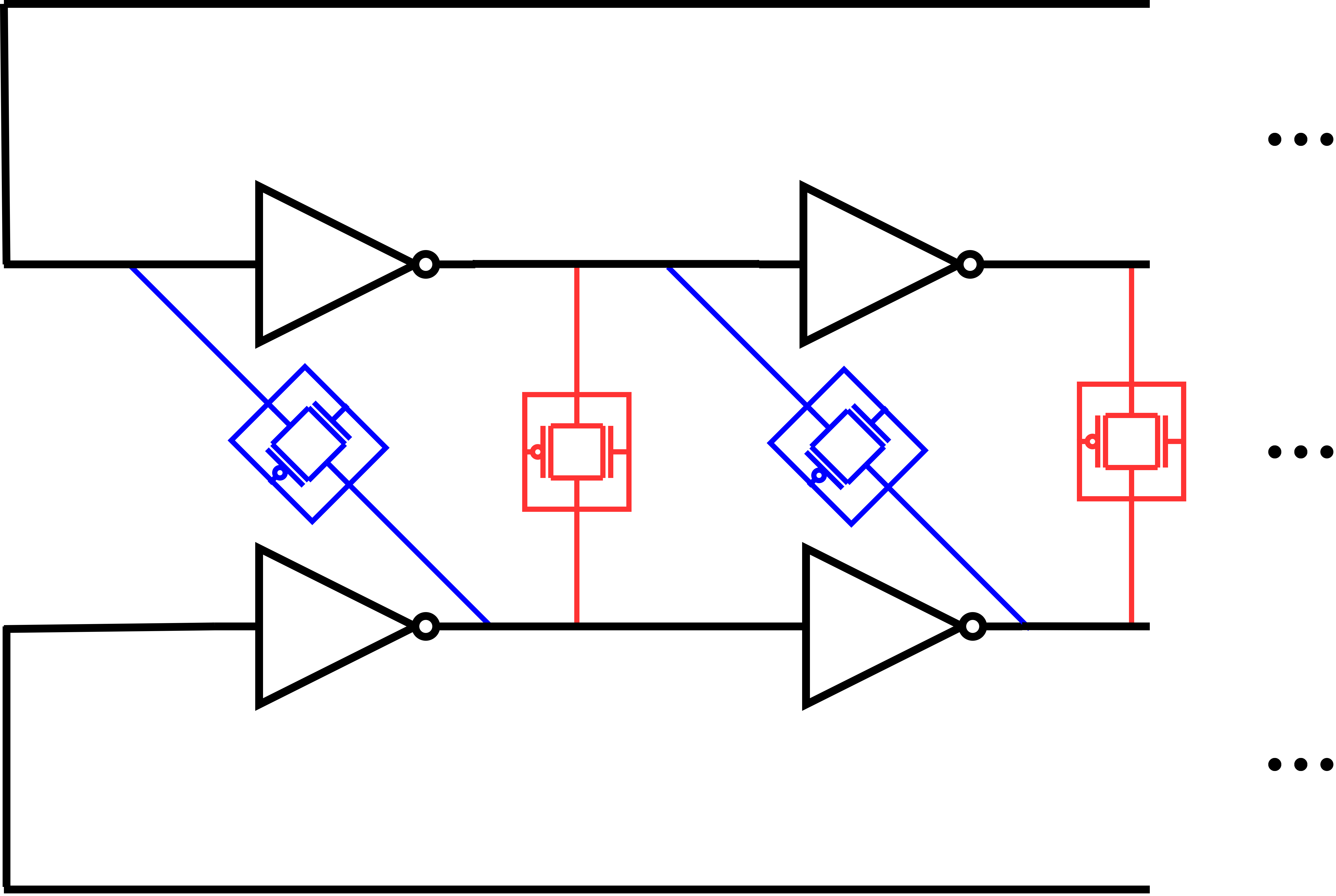}
    \caption{Schematic illustration of programmable weight polarity using CMOS transmission gates. Red paths realize positive coupling, and blue paths implement negative coupling by introducing a phase shift. Each oscillator pair uses this structure to encode signed interactions~\cite{lo2023ising}.}
    \label{fig:weight_circuit}
\end{figure}

In practice, both positive and negative coupling weights can be realized using CMOS transmission gates~\cite{moy20221,lo2023ising}. As illustrated in Figure~\ref{fig:weight_circuit}, each oscillator pair is connected through two distinct coupling paths: one for excitatory (positive) interaction and one for inhibitory (negative) interaction. These paths are selectively enabled based on the sign of the desired weight. Specifically, the red path activates in-phase (positive) coupling, while the blue path activates out-of-phase (negative) coupling by introducing a phase shift through an additional inverter stage. This dual-gate scheme enables compact, programmable, and polarity-selective coupling, directly supporting OscNet v1.5’s sparse weight matrices.

Sparsity plays a critical role in enabling efficient hardware implementation~\cite{dave2021hardware}. Compared to fully connected networks, sparse connectivity significantly reduces the number of required couplings, which are among the most area- and power-intensive components in analog designs. Sparse architectures minimize wire congestion, improve layout regularity, and reduce both static and dynamic power consumption. From a system design perspective, sparsity also aligns well with tile-based modular layouts, where each node interacts with only a small, fixed number of neighbors.

The structure and dynamics of OscNet v1.5 are naturally compatible with future \gls{VLSI} realizations. Its reliance on local interactions and forward-only computation makes it a promising candidate for low-power neuromorphic vision systems and edge inference hardware.

\section{Conclusion}
In this paper, we propose OscNet v1.5, a machine learning pipeline designed for CMOS oscillators. 
OscNet v1.5 is energy efficient image classification framework based on Hopfield Networks. 
OscNet v1.5 introduces sparse connectivity, combining biological plausibility with practical hardware efficiency. 
Experiments on MNIST show that OscNet v1.5 outperforms other biologically inspired models and even some deep learning baselines. 
Additionally, we demonstrate that its structure aligns well with CMOS-compatible oscillator arrays using \gls{SHIL}. 
In CMOS oscilltor implemention, OscNet v1.5 uses a small portion of connections but achieves almost the same performance as fully connected Hopfield Networks for image classification.
By using forward-only learning and using sparse connections, it achieves significant energy savings while maintaining strong classification performance.

\section*{Acknowledgment}
This work was generously supported by the Stanford-Samsung Research Initiative, the Stanford Graduate Fellowship, and the NSERC Fellowship from the Natural Sciences and Engineering Research Council of Canada, whose support is gratefully acknowledged.

% ICCC template
% {
%     \small
%     \bibliographystyle{ieeenat_fullname}
%     \bibliography{main}
% }

% arxiv
\bibliographystyle{unsrt}
\bibliography{main}

\begin{thebibliography}{10}

\bibitem{spatialbot}
Wenxiao Cai, Iaroslav Ponomarenko, Jianhao Yuan, Xiaoqi Li, Wankou Yang, Hao
  Dong, and Bo~Zhao.
\newblock Spatialbot: Precise spatial understanding with vision language
  models.
\newblock {\em arXiv preprint arXiv:2406.13642}, 2024.

\bibitem{ml_finance_1}
Matthew~F Dixon, Igor Halperin, and Paul Bilokon.
\newblock {\em Machine learning in finance}, volume 1170.
\newblock Springer, 2020.

\bibitem{ml_health_1}
Hafsa Habehh and Suril Gohel.
\newblock Machine learning in healthcare.
\newblock {\em Current genomics}, 22(4):291, 2021.

\bibitem{train_cost_1}
Sasha Luccioni, Yacine Jernite, and Emma Strubell.
\newblock Power hungry processing: Watts driving the cost of ai deployment?
\newblock In {\em The 2024 ACM Conference on Fairness, Accountability, and
  Transparency}, pages 85--99, 2024.

\bibitem{train_cost_2}
Nesrine Bannour, Sahar Ghannay, Aur{\'e}lie N{\'e}v{\'e}ol, and Anne-Laure
  Ligozat.
\newblock Evaluating the carbon footprint of nlp methods: a survey and analysis
  of existing tools.
\newblock In {\em Proceedings of the second workshop on simple and efficient
  natural language processing}, pages 11--21, 2021.

\bibitem{train_cost_3}
Andrew~A Chien, Liuzixuan Lin, Hai Nguyen, Varsha Rao, Tristan Sharma, and
  Rajini Wijayawardana.
\newblock Reducing the carbon impact of generative ai inference (today and in
  2035).
\newblock In {\em Proceedings of the 2nd workshop on sustainable computer
  systems}, pages 1--7, 2023.

\bibitem{train_cost_4}
Radosvet Desislavov, Fernando Mart{\'\i}nez-Plumed, and Jos{\'e}
  Hern{\'a}ndez-Orallo.
\newblock Compute and energy consumption trends in deep learning inference.
\newblock {\em arXiv preprint arXiv:2109.05472}, 2021.

\bibitem{graph_coloring}
Richelle~L. Smith and Thomas~H. Lee.
\newblock Polychronous oscillatory cellular neural networks for solving graph
  coloring problems.
\newblock {\em IEEE Open Journal of Circuits and Systems}, 4:156--164, 2023.

\bibitem{oscnet}
Wenxiao Cai, Zongru Li, and Thomas~H Lee.
\newblock Oscnet: Machine learning on cmos oscillator networks.
\newblock {\em IEEE Transactions on Circuits and Systems II: Express Briefs},
  2025.

\bibitem{anitha2021hyperbolic}
K~Anitha, R~Dhanalakshmi, K~Naresh, and D~Rukmani~Devi.
\newblock Hyperbolic hopfield neural networks for image classification in
  content-based image retrieval.
\newblock {\em International Journal of Wavelets, Multiresolution and
  Information Processing}, 19(01):2050059, 2021.

\bibitem{abernot2023training}
Madeleine Abernot and Aida Todri-Sanial.
\newblock Training energy-based single-layer hopfield and oscillatory networks
  with unsupervised and supervised algorithms for image classification.
\newblock {\em Neural Computing and Applications}, 35(25):18505--18518, 2023.

\bibitem{hebb1}
Paul Adams.
\newblock Hebb and darwin.
\newblock {\em Journal of theoretical Biology}, 195(4):419--438, 1998.

\bibitem{hebb2}
Stephen Grossberg.
\newblock Adaptive pattern classification and universal recoding: I. parallel
  development and coding of neural feature detectors.
\newblock {\em Biological cybernetics}, 23(3):121--134, 1976.

\bibitem{mnist}
Yann LeCun, L{\'e}on Bottou, Yoshua Bengio, and Patrick Haffner.
\newblock Gradient-based learning applied to document recognition.
\newblock {\em Proceedings of the IEEE}, 86(11):2278--2324, 1998.

\bibitem{hopfield}
Hubert Ramsauer, Bernhard Sch{\"a}fl, Johannes Lehner, Philipp Seidl, Michael
  Widrich, Thomas Adler, Lukas Gruber, Markus Holzleitner, Milena Pavlovi{\'c},
  Geir~Kjetil Sandve, et~al.
\newblock Hopfield networks is all you need.
\newblock {\em arXiv preprint arXiv:2008.02217}, 2020.

\bibitem{cheng1996application}
Kuo-Sheng Cheng, Jzau-Sheng Lin, and Chi-Wu Mao.
\newblock The application of competitive hopfield neural network to medical
  image segmentation.
\newblock {\em IEEE transactions on medical imaging}, 15(4):560--567, 1996.

\bibitem{oscillator_spiking1}
Eugene~M. Izhikevich and Frank~C. Hoppensteadt.
\newblock Polychronous wavefront computations.
\newblock {\em Int. J. Bifurc. Chaos}, 19:1733--1739, 2009.

\bibitem{isf_1}
Brian Hong and Ali Hajimiri.
\newblock A general theory of injection locking and pulling in electrical
  oscillators—part i: Time-synchronous modeling and injection waveform
  design.
\newblock {\em IEEE Journal of Solid-State Circuits}, 54:2109--2121, 2019.

\bibitem{kuramoto_eq}
Hidetsugu Sakaguchi, Shigeru Shinomoto, and Yoshiki Kuramoto.
\newblock Local and grobal self-entrainments in oscillator lattices.
\newblock {\em Progress of Theoretical Physics}, 77:1005--1010, 1987.

\bibitem{potts_1}
F.~Y. Wu.
\newblock The potts model.
\newblock {\em Reviews of Modern Physics}, 54:235--268, 1982.

\bibitem{wang2024shem}
Yu-Neng Wang and Sara Achour.
\newblock Shem: A hardware-aware optimization framework for analog computing
  systems.
\newblock {\em arXiv preprint arXiv:2411.03557}, 2024.

\bibitem{mohseni2022ising}
Naeimeh Mohseni, Peter~L McMahon, and Tim Byrnes.
\newblock Ising machines as hardware solvers of combinatorial optimization
  problems.
\newblock {\em Nature Reviews Physics}, 4(6):363--379, 2022.

\bibitem{belyaev2020classification}
MA~Belyaev and AA~Velichko.
\newblock Classification of handwritten digits using the hopfield network.
\newblock In {\em IOP conference series: materials science and engineering},
  volume 862, page 052048. IOP Publishing, 2020.

\bibitem{moy20221}
William Moy, Ibrahim Ahmed, Po-wei Chiu, John Moy, Sachin~S Sapatnekar, and
  Chris~H Kim.
\newblock A 1,968-node coupled ring oscillator circuit for combinatorial
  optimization problem solving.
\newblock {\em Nature Electronics}, 5(5):310--317, 2022.

\bibitem{lo2023ising}
Hao Lo, William Moy, Hanzhao Yu, Sachin Sapatnekar, and Chris~H Kim.
\newblock An ising solver chip based on coupled ring oscillators with a 48-node
  all-to-all connected array architecture.
\newblock {\em Nature Electronics}, 6(10):771--778, 2023.

\bibitem{dave2021hardware}
Shail Dave, Riyadh Baghdadi, Tony Nowatzki, Sasikanth Avancha, Aviral
  Shrivastava, and Baoxin Li.
\newblock Hardware acceleration of sparse and irregular tensor computations of
  ml models: A survey and insights.
\newblock {\em Proceedings of the IEEE}, 109(10):1706--1752, 2021.

\end{thebibliography}

\end{document}